\documentclass[journal]{IEEEtran}

\usepackage{cite}

\ifCLASSINFOpdf
  \usepackage[pdftex]{graphicx}
  \graphicspath{{fig/}}
  \DeclareGraphicsExtensions{.pdf,.jpeg,.png}
\else
\fi

\usepackage{amsmath,amsfonts,amssymb}
\usepackage{algpseudocode}
\usepackage{algorithm}
\usepackage{array}
\usepackage{multirow}

\usepackage[ansinew]{inputenc}
\usepackage{bm}
\usepackage{subcaption}	
\usepackage{import}
\usepackage{xcolor}
\usepackage[squaren]{SIunits}
\usepackage{upgreek}
\usepackage{hyperref}
\usepackage{booktabs}

\newcolumntype{C}[1]{>{\centering\arraybackslash}p{#1}}

\begin{document}
\renewcommand\arraystretch{1.4}
\title{Improved Exploring Starts by Kernel Density Estimation-Based State-Space Coverage Acceleration in Reinforcement Learning}
\author{\IEEEauthorblockN{Maximilian Schenke, Oliver Wallscheid}
\thanks{M. Schenke is with the Department of Power Electronics and Electrical Drives and O. Wallscheid is with the Department of Automatic Control, both at Paderborn University, Germany. \mbox{E-mail:} \{schenke, wallscheid\}@lea.uni-paderborn.de}}

\maketitle
\begin{abstract}
Reinforcement learning (RL) is currently a popular research topic in control engineering and has the potential to make its way to industrial and commercial applications. Corresponding RL controllers are trained in direct interaction with the controlled system, rendering them data-driven and performance-oriented solutions. The best practice of exploring starts (ES) is used by default to support the learning process via randomly picked initial states. However, this method might deliver strongly biased results if the system's dynamic and constraints lead to unfavorable sample distributions in the state space (e.g., condensed sample accumulation in certain state-space areas). To overcome this issue, a kernel density estimation-based state-space coverage acceleration (DESSCA) is proposed, which improves the ES concept by prioritizing infrequently visited states for a more balanced coverage of the state space during training. Compared to neighbouring methods in the field of count-based exploration, DESSCA can also be applied to continuous state spaces without the need for artificial discretization of the states. Moreover, the algorithm allows to define arbitrary reference state distributions such that the state coverage can be shaped w.r.t. the application needs. Considered test scenarios are mountain car, cartpole and electric motor control environments. Using DQN and DDPG as exemplary RL algorithms, it can be shown that DESSCA is a simple yet effective algorithmic extension to the established ES approach that enables an increase in learning stability as well as the final control performance.   
\end{abstract}


\IEEEpeerreviewmaketitle

\section{Introduction}

\IEEEPARstart{T}{he} utilization of reinforcement learning (RL) controllers in industrial control applications is gradually becoming a reasonable alternative to conventional control approaches. The data-driven nature of RL algorithms allows the controller to tune itself in direct interaction with the controlled system, which enables the learning of sophisticated control sequences without the necessity of a plant model. This advantage comes at the cost of training time, which must be invested such that the RL controller (the agent) becomes well and thoroughly acquainted with the dynamics of the controlled system (the environment). In order to make the most out of the given training time, many modern RL algorithms include features like importance sampling \cite{shelton2001importance}, experience replay buffers \cite{fedus2020revisiting} and / or internal model learning \cite{moerland2021modelbased}, which all pursue the goal of sampling efficiency. 

A very intuitive and established approach is the utilization of exploring starts (ES) in RL algorithms. When applicable, an initial state is selected at random such that the RL controller can gather information about different regions of the state space even before it has learned to maneuver them safely. This best practice has proven its value in many different scenarios \cite{DBLP:journals/corr/abs-2002-03585}, but whenever the selection of arbitrary initial states is possible, this degree of freedom may also permit utilization that is more expedient than random sampling \cite{DBLP:journals/corr/abs-2012-00724}. 

\begin{figure}[htp]
\centering
\includegraphics[scale=0.42]{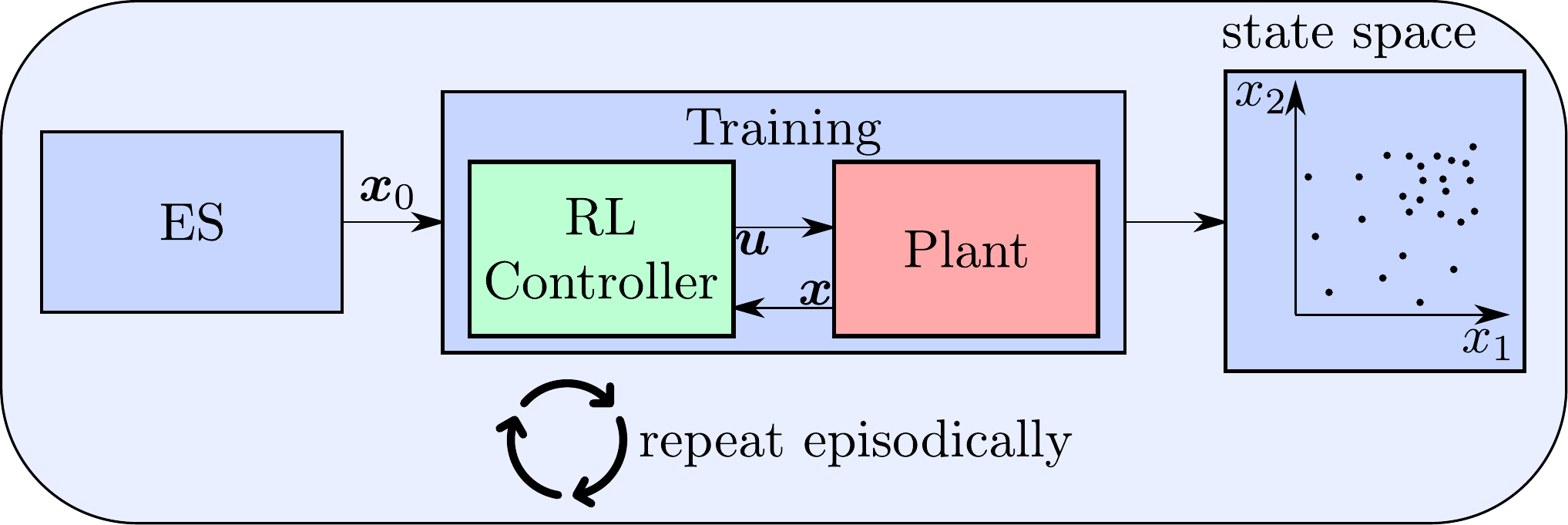}
\caption{Conceptual depiction of a biased state distribution when using random exploring starts (ES) to determine an initial state of the plant system during episodic training}
\label{fig:es_concept}
\end{figure}

The internal system dynamics will naturally lead to sample accumulation, e.g., in the proximity of attractive equilibria, or vice versa to an insufficient coverage around state limitations in the context of safety shutdowns. This behavior, conceptually depicted in Fig. \ref{fig:es_concept}, is attribute to the plant system and the corresponding control task, it can never be entirely compensated by an initialization strategy. Nonetheless, a target-oriented initialization strategy can be utilized to balance the state-space exploration to some extent, enabling a more efficient usage of training time, increased controller robustness and improved performance.

A brief outline on RL exploration is drawn in Sec. \ref{sec:theory}, followed by the proposed density estimation-based extension to exploring starts in Sec. \ref{sec:DESSCA}. The presented algorithm is tested and evaluated in Sec. \ref{sec:verification} and \ref{sec:result}. Lastly, possible extensions and open questions are reviewed in Sec. \ref{sec:outlook}.

\section{Background}
\label{sec:theory}
When using standard ES in RL, initial states are sampled from a random distribution which is oftentimes simply a uniform distribution of the (feasible) state space. Due to the dynamics of a plant system being unknown and the behavior of the learning control agent being adapted at run time, no general predictions about the course of the resulting state trajectory can be made. Hence, analytical trajectory planning is not an option and alternatives are required. Here, collected information about already visited states can be utilized to estimate which regions of the state space were overrepresented or underrepresented in the past training. This information can then be exploited to select the next initial state in the region of sparsest coverage and, therefore, make optimal use of exploration potential.

The utilization of any initialization strategy, no matter how sophisticated, firstly assumes the possibility to freely choose the initial state of the plant system. This can always be assured in simulated environments, however, real-world systems are more complicated and time consuming to initialize. In the scope of this contribution it is assumed that a low-performance controller (LPC) must be available to steer a real-world plant system to the intended initial state, which is given for usual industrial applications. Using the LPC within the context of the initialization strategy allows to replace it with a better-performing RL controller that is more reliable after the training. Particularly, the LPC does not need to fulfill special demands concerning the transient performance and even stationary errors (which lead to unintended initial states) can be handled to some degree as long as their appearance is fed back to the initialization strategy (which is not the case in ES).

This section gives a short overview about the application of RL methods in the context of automatic control before reviewing established exploration methods  and outlining the theoretical foundations for the suggested approach. 
\subsection{Reinforcement Learning in Control}
In the context of this paper, RL is utilized to solve the well-known optimal control problem
\begin{align}
\begin{split}
    &\underset{\bm{u}}{\text{max}} \; \sum_{i=k}^\infty \gamma ^{i-k} r(\bm{x}_k, \bm{u}_k),
    \\
    \text{s.t.} \quad & \bm{x}_{k+1}=\bm{f}(\bm{x}_k, \bm{u}_k),
    \\
    & \bm{h}(\bm{x}_k, \bm{u}_k) = \bm{0},
    \\
    & \bm{l}(\bm{x}_k,\bm{u}_k) \leq \bm{0},
    \label{eq:optimal_control}
\end{split}
\end{align}
with the reward function $r$, the system's state vector $\bm{x}$, the control input $\bm{u}$ and the sampling index $k$. The discount factor $\gamma \in ]0,1[$ allows to prioritize the control performance on a short-sighted or far-sighted time horizon. Please note that the optimal control problem is mostly formulated as a cost minimization problem, whereas the RL community usually considers the reward maximization task. This contribution will make use of the latter definition. The control plant is furthermore subject to the system dynamics $\bm{f}$ and to different limitations described by equality constraints $\bm{h}$ or inequality constraints $\bm{l}$. It is assumed that violating $\bm{h}$ or $\bm{l}$ will lead to an emergency shutdown (e.g., through superimposed safety mechanisms) which terminates the running episode and causes a plant system restart. 

Whereas conventional optimal control methods make use of an approximate system model $\hat{\bm{f}}$ in order to compute the best possible input trajectory \cite{rawlings2017model}, RL is capable to provide optimal control performance without the need for a plant model. Influencing the plant system's dynamic behavior is then learned during the training phase of the RL controller by direct interaction within a closed loop. Simplifications of the control problem (\ref{eq:optimal_control}), as often applied in optimal control algorithms like model predictive control to render the optimization task convex and, hence, solvable for standard optimizers, are not needed.

The maximization task in (\ref{eq:optimal_control}) is usually viewed as a sequence of one-step problems as defined by the Bellman optimality equation
\begin{align}
\begin{split}
    V^*(\bm{x}_k)&=\underset{\bm{u}_k}{\text{max}} \; \mathbb{E}\left\{ R_{k+1}
    + \gamma V^*(\bm{X}_{k+1}) | \bm{X}_k=\bm{x}_k\right\},
    \label{eq:optimal_v}
\end{split}
\end{align}
in terms of the value function $V$ or
\begin{align}
\begin{split}
    Q^*(\bm{x}_k, \bm{u}_k^*)&=\underset{\bm{u}_k}{\text{max}} \; \mathbb{E}\left\{ R_{k+1} \right. \\
    & \quad \left.
    + \gamma Q^*(\bm{X}_{k+1}, \bm{U}_{k+1}) | \bm{X}_k=\bm{x}_k, \bm{U}_k=\bm{u}_k \right\},
    \label{eq:optimal_q}
\end{split}
\end{align}
in terms of the action value function $Q$. Capital letters $R, \bm{X}, \bm{U}$ denote random variables in this section and the expected value is denoted by $\mathbb{E}\{\cdot\}$.

Given this perspective, the control agent is capable of optimal performance by consecutively selecting the best available control signal $\bm{u}$ that maximizes the right-hand side of either Bellman equation. Therefore, $V$ or $Q$ must firstly be learned for each state and for the underlying control policy, motivating the consideration of state-space exploration methods such as presented in this contribution.

Contrary to usual controller application, the training phase of RL controllers is often conducted in an episodic manner, which means that reinitialization of the plant is an inherent component of the training. This has several reasons:
\begin{itemize}
    \item Training with time-limited episodes enables the usage of an initialization strategy that supports the learning process.
    \item Emergency plant shutdowns may be required to prevent harm (especially in the early training) which necessitate the reinitialization of the control loop and, hence, end one episode to start the next.
    \item The control problem might be repetitive, which is often the case in, e.g., production processes including industrial robots.
    \item The control problem might be terminating naturally after a certain time limit or when a target state has been reached, e.g., an aircraft after landing.
\end{itemize}
Therefore, it is assumed in the following that the closed-loop control system is regularly subjected to episodic restarts, which will be actively used as part of the exploration process.

\subsection{Exploration in RL}
\label{sec:exploration}
The paradigm of exploration and exploitation is the foundation for most RL algorithms. Herein, exploration describes the discovery of the state and action space that is available in the given control environment in order to determine $V$ or $Q$ accurately. Exploitation is the application of the input signal that is considered to be optimal in terms of $V$ or $Q$. While exploitation is clearly defined via the Bellman optimality equation (\ref{eq:optimal_v}), (\ref{eq:optimal_q}) there are many valid approaches to define exploration mechanisms with different advantages and disadvantages. 

Three main classes of exploration strategies can be distinguished according to \cite{mcfarlane2018survey}:
\begin{itemize}
    \item[a)] Undirected exploration, which makes use of random action selection to maneuver the state space. The well-known $\epsilon$-greedy method \cite{DBLP:journals/corr/MnihKSGAWR13} falls under this category.
    \item[b)] Directed exploration based on the learning process, which contains mainly count-based approaches that target a balanced number of visits to each state \cite{tang2017exploration}.
    \item[c)] Directed exploration based on alternative learning models, where approaches are based, e.g., on concepts of learned system models \cite{606886}, which allows trajectory planning, or built-in curiosity \cite{still2012information}, which encourages an RL agent to explore what has not been seen.
\end{itemize}
Further options may be considered according to \cite{garcia2015comprehensive} if expert knowledge is available about the task at hand. Corresponding methods may include the usage of advice given by a teacher entity that communicates with the control agent and can therefore influence the agent's decisions, or the control agent can be initialized such that impracticable state-space regions are avoided from the start.

The approach that is presented in the following can be classified in category (b) of the above mentioned strategies. This means that the exploration strategy makes use of the data collected during training, and that no expert knowledge (such as a plant model) is used in the process. A review of existing approaches within this category will be delivered in the next subsection.

\subsection{Count-Based Exploration: State of the Art}
The idea behind count-based exploration is rather intuitive, but unfortunately it is not straightforward to convey it to the continuous state domain, which this contribution is addressing. In problems with a discrete state space, count-based exploration can be easily employed by maintaining a table of counters for each accessible state \cite{Thrun92efficientexploration}. This table then allows to assess which states are rarely visited, such that this information can be used to explore what is mostly unknown. 

Approaching the counting problem by using a density model is not new. Similar ideas have been presented, e.g., in \cite{bellemare2016unifying} and \cite{ostrovski2017countbased}. Within these studies, a large-scale but discrete (and therefore still countable) state space is accounted for with a density model to overcome the challenge of many states being sparsely visited. The utilization of a Gaussian mixture model (GMM) as density estimator in \cite{zhao2020curiositydriven} allowed the extension of corresponding methods to continuous state spaces. In contrast to the later discussed kernel density estimation (KDE) the utilization of GMMs allows a faster but less flexible density estimate because the number of compounded distributions needs to be set before starting the learning process and because their form is - obviously - strictly Gaussian. 

In \cite{tang2017exploration} the counting problem has been resolved by using a locality-sensitive hashing algorithm to map a continuous state vector to a finite hash table, which can be understood as a discretization approach. A hashing algorithm accepts data of arbitrary length and transforms it into a well-defined format, the so-called hash code, which is in this case a table index. The special case of locality-sensitive hashing additionally conserves similarities, meaning that similar state inputs point to neighbouring or even identical table entries. Naturally, the granularity of this approach to discretization determines the loss of information. Therefore, a high resolution can only be achieved by storing significantly large tables. Further, hashing methods are usually not invertible, which means that a rarely visited state vector cannot be reconstructed from the hash table, rendering the user unable to utilize the state-space coverage within ES.

In both approaches, the ones with a density model as well as the ones with hashing algorithms, the information about the number of state visits is used as an augmentation to the reward function
\begin{align}
    \tilde{r}_{k+1}&=r_{k+1}+\frac{\beta}{\sqrt{n(\bm{x}_{k+1})}},
\end{align}
wherein $n(\bm{x}_{k+1})$ is a (pseudo-)count of the occurrence of $\bm{x}_{k+1}$ and $\beta$ is a scaling factor. This way, exploring behavior is interpreted as part of the optimal policy, forcing the agent to learn how to explore its environment as an inherent part of its control strategy. This leads to a bias of the originally intended optimal control problem which conveys to a suboptimal control policy. Such a bias is hardly acceptable within the context of controlling technical processes. Therefore, the proposed method does not need to alter the actual reward function to ensure sufficient exploration. This is especially beneficial for applications with a interpretable reward signal (e.g., negative mean-squared control error of some physical quantity) since $V$ and $Q$ can then also be interpreted accordingly. 

\subsection{Contribution}
Instead of distorting the optimization task by modifying the reward, only the degree of freedom that is available through the initial state is to be utilized using a density estimation-based approach for continuous state counting. This way, the resulting policy remains untouched from the exploration strategy and the occurrence of diverse states is more balanced than with randomly selected ES. This approach to directed, model-free exploration within continuous state-spaces is the main contribution of this article. Additionally, reference state distributions can be freely defined such that application-specific state coverage strategies can be pursued.

\section{Proposed Approach: Density Estimation-Based State-Space Coverage Acceleration (DESSCA)}
\label{sec:DESSCA}
\subsection{Directed Exploration for Continuous State Spaces}
\label{sec:theory_c}
\begin{figure*}[htb]
\begin{subfigure}[c]{0.33\textwidth}
\includegraphics[width=1\textwidth]{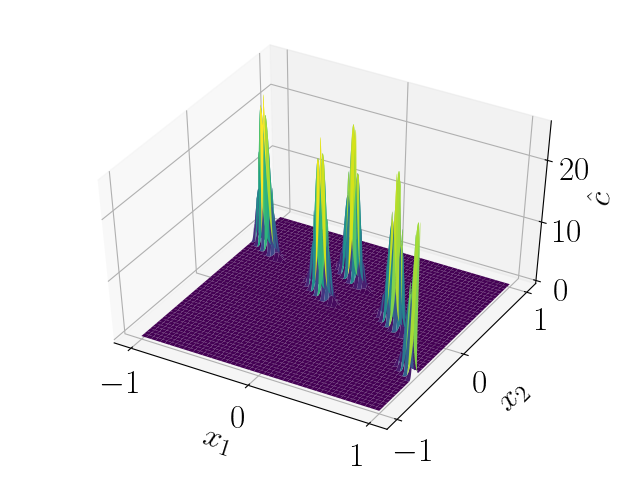}
\subcaption{$b=0.1$}
\label{fig:bw_01}
\end{subfigure}
\begin{subfigure}[c]{0.33\textwidth}
\includegraphics[width=1\textwidth]{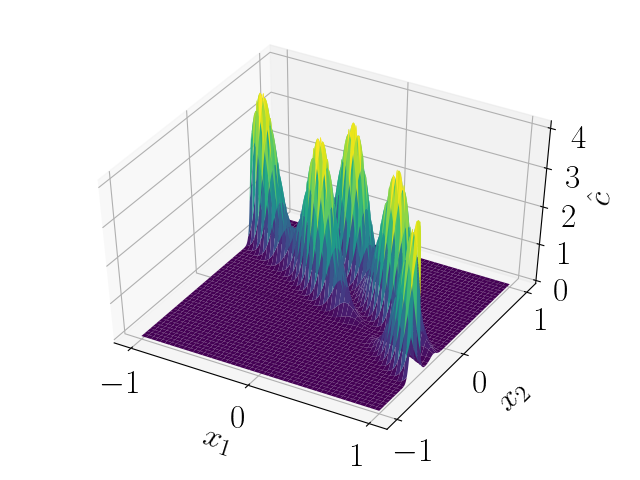}
\subcaption{$b=0.25$}
\label{fig:bw_025}
\end{subfigure}
\begin{subfigure}[c]{0.33\textwidth}
\includegraphics[width=1\textwidth]{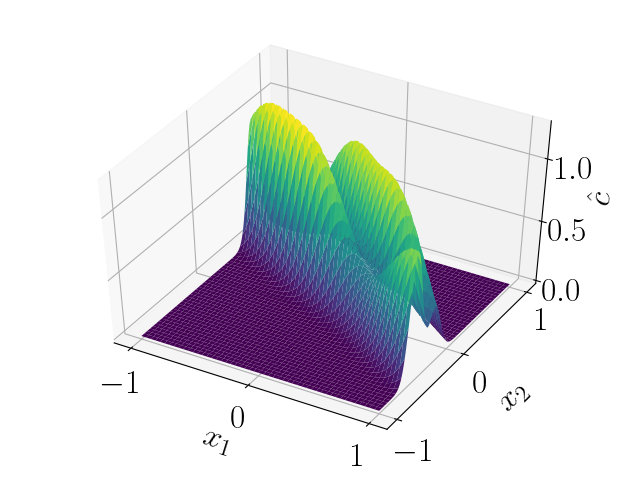}
\subcaption{$b=0.5$}
\label{fig:bw_05}
\end{subfigure}
\caption{Exemplary coverage density estimation $\hat{c}(\bm{x})$ of the same sampling set $\mathcal{X}_k$ (with $|\mathcal{X}_k|=5$) for different selections of the bandwidth parameter $b$ and Gaussian kernel $g(\cdot)$ on a normalized, two-dimensional state space ($x_i \in [-1, 1]$)}
\label{fig:bw}
\end{figure*}

For the distinction of the highly visited and rather unvisited state-space regions, the coverage density function $c(\bm{x})$ is introduced. This function will replace the counter table that is usually used in count-based approaches in discrete state spaces. Following the idea of a probability density function, the coverage density maps the continuous state vector $\bm{x}$ to a corresponding probability density of occurrence. As $c(\bm{x})$ is subject to the dynamics of the unknown plant system, the influence of the adaptive and learning RL controller and the selected initialization strategy it is not an option to derive an analytical closed form solution for $c(\bm{x})$. However, with the use of a set of past visited states $\mathcal{X}_k$, which have been visited until time step $k$, it is possible to yield an estimated coverage density $\hat{c}(\bm{x})$ using KDE:
\begin{align}
    \hat{c}(\bm{x})=\frac{1}{|\mathcal{X}_k|b}\sum_{\bm{x}_i \in \mathcal{X}_k}g\left(
    \frac{\bm{x} - \bm{x}_i}{b}
    \right),
\end{align}
with the kernel function $g(\cdot)$ and the bandwidth parameter \mbox{$b>0$}. The bandwidth can be interpreted as the reciprocal resolution of the KDE procedure. A smaller bandwidth corresponds to a higher resolution whereas a higher bandwidth leads to lower resolution as depicted in Fig. \ref{fig:bw}. The selection of the optimal bandwidth is problem dependent. Especially in the context of RL it is important to match $b$ to the generalizing capability of the control agent and the environment.

The least represented state from the state-space is then given by the global minimum of $\hat{c}(\bm{x})$. Typical candidates for $g(\cdot)$ are, e.g., the rectangular, the Epanechnikov or the Gaussian kernel \cite{wkeglarczyk2018kernel}. The utilization of $\mathcal{X}_k$ within the formulation of this estimator justifies the affiliation of the featured method to the category of directed exploration methods that are based on the data acquired during the learning process (Sec. \ref{sec:exploration}).

\begin{figure}[htp]
\centering
\includegraphics[scale=0.42]{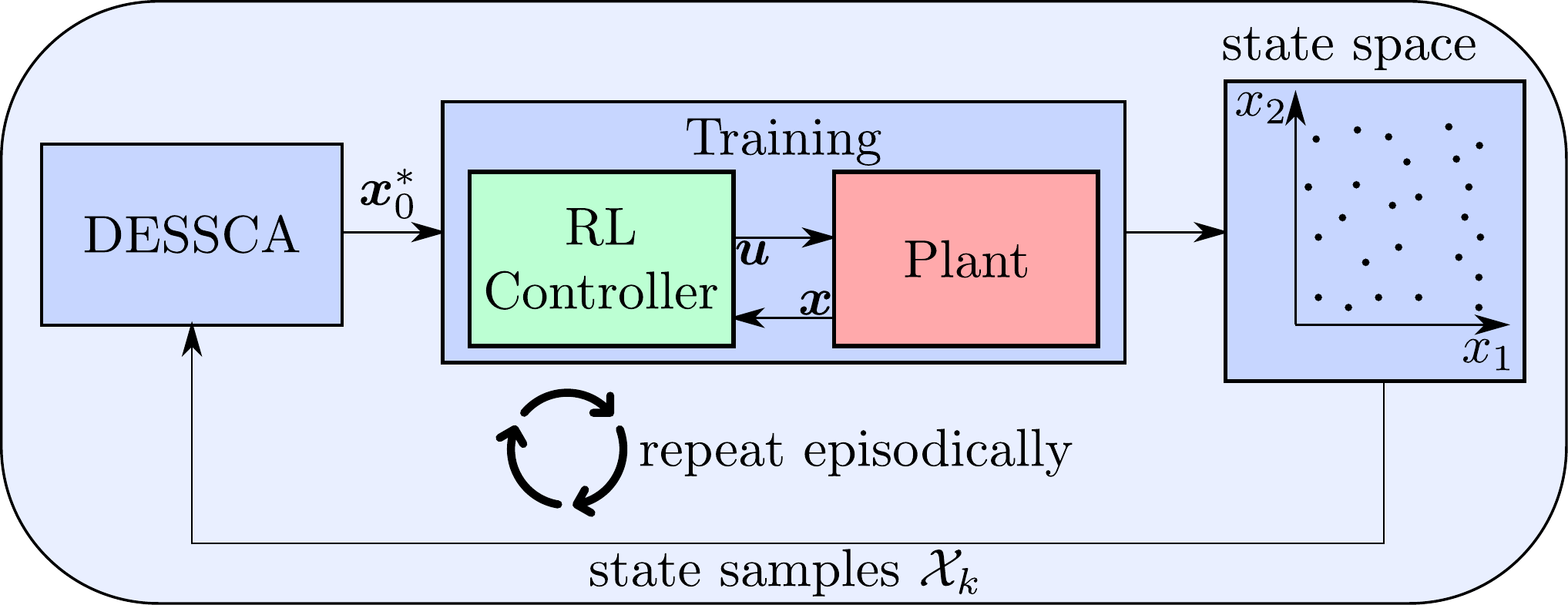}
\caption{Conceptual depiction of a less biased state distribution when using DESSCA to determine an initial state of the plant system during episodic training}
\label{fig:dessca_concept}
\end{figure}

The reference coverage density function $c^*(\bm{x})$ is to be set by the user and can therefore incorporate knowledge about the prioritized operation region of the given control application. It furthermore allows the consideration of state-space limitations and the exclusion of irrelevant operation regions. If no such knowledge is available a priori, the choice of balanced exploration is a natural approach, leading to $c^*(\bm{x})$ being a uniform distribution. The difference of both coverage functions 
\begin{align}
e_c(\bm{x}) := c^*(\bm{x})-\hat{c}(\bm{x}),
\end{align}
is considered as the exploration metric that is to be maximized before each exploring start to find an optimal starting state $\bm{x}_0^*$:
\begin{align}
\label{eq:DESSCA_cost_fct}
    \bm{x}_0^*:=\underset{\bm{x}}{\text{arg}\,\text{max}} \; e_c(\bm{x}).
\end{align}
Please note that the difference coverage $e_c$ can not be interpreted as a probability density. The described procedure can be labeled as density estimation-based state-space coverage acceleration (DESSCA), which is summarized in Alg. \ref{alg:dessca}. A schematic of the aspired behavior is given in Fig. \ref{fig:dessca_concept}. In comparison to Fig. \ref{fig:es_concept} the acquisition of a more balanced state history $\mathcal{X}_k$ allows a better learning behavior.

{\linespread{1.3}
\begin{algorithm}
\caption{DESSCA pseudocode}
\label{alg:dessca}
\begin{algorithmic}
\State \textbf{input:} Reference coverage density function $c^*(\bm{x})$, kernel function $g(\cdot)$, bandwidth $b$
\State \textbf{output:} RL training with optimal episode initialization $\bm{x}_0^*$
\While{Learning}
\If {$|\mathcal{X}_k| > 0$}
    \State Estimate observed coverage density via KDE:
    \State \qquad $\hat{c}(\bm{x})=\frac{1}{|\mathcal{X}_k|b}\sum_{\bm{x}_i \in \mathcal{X}_k}g\left(
    \frac{\bm{x} - \bm{x}_i}{b}
    \right)$
    \State Maximize estimated difference coverage density: 
    \State \qquad $\bm{x}_0^*=\underset{\bm{x}}{\text{arg}\,\text{max}} \; c^*(\bm{x})-\hat{c}(\bm{x})$
\Else 
    \State Maximize $c^*$: $\bm{x}_0^*=\underset{\bm{x}}{\text{arg}\,\text{max}} \; c^*(\bm{x})$
\EndIf
\State Train for one episode with initial state $\bm{x}_0^*$
\State Memorize newly visited states in $\mathcal{X}_k$
\EndWhile
\end{algorithmic}
\end{algorithm}
}

\subsection{Implementation}
\label{subsec:realization}
This section outlines an implementation of the proposed DESSCA algorithm. For the verification presented in Sec.~\ref{sec:verification}, the algorithm has been realized in Python version 3.7 \cite{10.5555/1593511}, using the KDE library kalepy \cite{Kelley2021}. For the following investigations the KDE is configured with a bandwidth of $b=0.1$ (cf., Fig. \ref{fig:bw_01}). Each dimension of the state vector is normalized to the range $x_i \in [-1,1]$ and a Gaussian kernel is selected by default. The analyzed applications are detailed in their corresponding subsections, the utilized deep learning frameworks are kerasRL \cite{plappert2016kerasrl}, kerasRL2 \cite{McNally2019} and tensorflow \cite{tensorflow2015-whitepaper}.

Comparing multidimensional probability density functions based on a discrete number of samples, it is likely that \eqref{eq:DESSCA_cost_fct} becomes a nonlinear and also multimodal optimization problem which requires a global optimization approach. In the context of control engineering and technical applications it is sensible to assume that the state space of any plant system is limited by constraints (e.g., upper and lower bounds) that correspond to safety- and performance-oriented criteria. These constraints allow to narrow down the search space on which the optimization needs to be performed and, thus, enable the utilization of several swarm-based heuristic opimizers. The proposed DESSCA algorithm will therefore make use of particle swarm optimization (PSO) \cite{kennedy1995particle} to find ${\text{arg}\,\text{max}}_{\bm{x}} \; e_c(\bm{x})$, but many other heuristic approaches, e.g., evolutionary algorithms or simulated annealing \cite{talbi2009metaheuristics} may be considered instead. For this specific implementation the Python library PySwarms is used to provide the PSO method \cite{pyswarmsJOSS2018}, and its parameterization is provided in Tab. \ref{tab:pso_params}.

\begin{table}[htb]
\centering
\begin{tabular}{l rl}
\toprule
\textbf{description} & \textbf{value} & \\
\hline
number of particles & $10 \cdot \text{dim}(\bm{x})$\\
number of iterations & $10 \cdot \text{dim}(\bm{x})$ + 5\\
cognitive parameter & $2$\\
social parameter   & $2$\\
particle inertia    & $0.6$\\
\bottomrule
\end{tabular}
\caption{Parameterization of PSO provided by PySwarms \cite{pyswarmsJOSS2018}}
\label{tab:pso_params}
\end{table}

The incorporation of feasible initial states is possible for both, default ES as well as DESSCA. The sensible state space can often be narrowed down to a region in which the plant system is controllable or at least stabilizable. The proper consideration of such conditions is enabled in the DESSCA algorithm by designing the reference coverage $c^*(\bm{x})$ with respect to the given limitations. 

Lastly, in some scenarios it may not be expedient to take the whole state history $\mathcal{X}_k$ into account for assessing the next initial state. As the RL controller is changing over the course of the training phase it may unlearn to maneuver state regions that were visited a long time ago. This can be accounted for by specifying $\mathcal{X}_k$ as a ring buffer, i.e., the total number of memorized states is limited to the most recent ones and, therefore, states can be revisited as soon as they have been forgotten. If combined with a replay buffer using RL algorithm, DESSCA can also utilize the same experience storage limiting its memory demand.

The presented DESSCA algorithm is available as an open-source code repository. It can be accessed via \cite{SupplementaryMaterial2021}.

\section{Experimental Tests}
\label{sec:verification}
To verify the proposed method, DESSCA is to be tested in different control scenarios. The general setup of this verification routine will be similar in all three considered environments: a total of 50 distinct RL control agents are trained using 1) the standard ES strategy with uniformly distributed initial states or 2) the proposed DESSCA algorithm as their initialization strategy. After the training phase the agents with either initialization strategy are subjected to a quantitative validation routine to determine whether the employed initialization strategy does lead to a meaningful difference. 

The scenarios under investigation are a) the mountain car, b) the cartpole and c) the current control of an electric motor. Herein, a) is an optimal control problem that terminates upon reaching a targeted state, preferably with lowest effort. The validation procedure consists of randomly initialized episodes that are to be solved by the agent. b) can be considered a pseudo tracking control problem, because a targeted state is to be reached and maintained but is never changed. An optimal controller should be able to hold the targeted state as long as possible without violating the plant's operation constraints. Validation is, again, performed over the course of many randomly initialized episodes. c) is a tracking control problem that includes a dynamically changing reference trajectory. Optimally, the controller assures that reference and plant state are equal while respecting the system's operation limits. Here, only one validation episode of extensive length is conducted, which includes a sequential traversing through many relevant operation points. 

The length of training and validation can be found in \mbox{Tab. \ref{tab:training_config}}. For comparability it has been assured that the validation episodes' demands for the ES and the DESSCA agents are identical. Detailed control problem specifications can be found in the appendix. The considered applications are also available as code examples at \cite{SupplementaryMaterial2021}. 

\begin{table}[htb]
\centering
\begin{tabular}{l r r r}
\toprule
& \textbf{mountain car} & \textbf{cartpole} & \textbf{PMSM} \\
\hline
total training steps           & $30\,000$ & $100\,000$ & $400\,000$\\
training steps per episode   &     $200$ &      $200$ &      $100$\\
validation episodes       &  $1\,000$ &   $1\,000$ &        $1$\\
validation steps per episode &     $200$ &      $200$ & $190\,500$\\
\bottomrule
\end{tabular}
\caption{Training and validation configuration for the three considered environments}
\label{tab:training_config}
\end{table}

 The performance is evaluated in terms of the theoretical maximum return $g_\text{max}$:
\begin{align}
    \label{eq:performance_normal}
    \frac{g}{g_\text{max}}&=\frac{\sum_k^K r_k}{\sum_k^K r_\text{max}}.
\end{align}
Wherein $r_\text{max}$ is the upper limit of the employed reward range. Reaching it at each time step can not be expected practically. Please note that this normalized quantity can be negative, depending on the reward specification.

\subsection{Mountain Car}
The mountain car environment from OpenAI Gym \cite{openaigym} is an optimal control problem in which a car needs to build up speed in order to escape a valley. In this scenario the goal never changes. The measured state space features the position and the speed of the car, the action signal is continuous and represents the acceleration of the car, which is insufficient to climb the mountains on its own. Therefore, the optimal control strategy features a swing-up maneuver using the slopes on both sides of the valley. Further information about the mountain car problem are given in the appendix in (\ref{eq:MC}).

As of Fig. \ref{fig:MC_Boxplots}, it can be seen that even for ES, the validation results are quite close to the theoretical maximum return, which means that only few room for improvement is left. Still, DESSCA manages to achieve a slightly better median performance and also the corresponding interquartile range is in favour of higher control quality, hinting at a more reliable and robust training process.

\subsection{Cartpole}
The cartpole is a classical control problem that is often used as a toy example to demonstrate the quality of nonlinear control approaches. In this scenario, a pendulum that is mounted on a cart is to be balanced in the upper equilibrium by accelerating the cart either to the left or to the right. The standard formulation of of this problem in the RL domain is, again, provided by OpenAI Gym \cite{openaigym}. A normalized system definition with further information is provided in the appendix in (\ref{eq:CP}).

The observed performance depicted in Fig. \ref{fig:CP_Boxplots} hints that a meaningful impact of the initialization strategy can only be concluded with benevolence. The median performance is worse using the DESSCA algorithm, but the interquartile range seems too similar to infer a striking difference. This observation indicates that the prioritization of underrepresented states is not particularly advantageous within the cartpole scenario, hinting that most relevant states can be visited naturally.

\subsection{Permanent Magnet Synchronous Motor Current Control}
The current control of electrical drives is a control problem of industrial relevance that can be solved using RL, as was shown in the publications \cite{8877848}, \cite{9376968}. The corresponding simulation environment is provided by the gym-electric-motor Python library \cite{Balakrishna2021}. In contrast to the mountain car and the cartpole environment, this control plant features a reference vector, rendering it an optimal tracking control problem. In the context of ES, and under the assumption of piecewise constant reference values, the reference is an interesting degree of freedom the initialization strategy may make use of to force exploration in all state variables. A more detailed plant specification can be found in the appendix. For this environment the memory buffer size of DESSCA was limited to $\mathcal{X}_k \leq 100\,000$.

\begin{figure*}[htb]
\begin{subfigure}[c]{0.33\textwidth}
\includegraphics[width=1\textwidth]{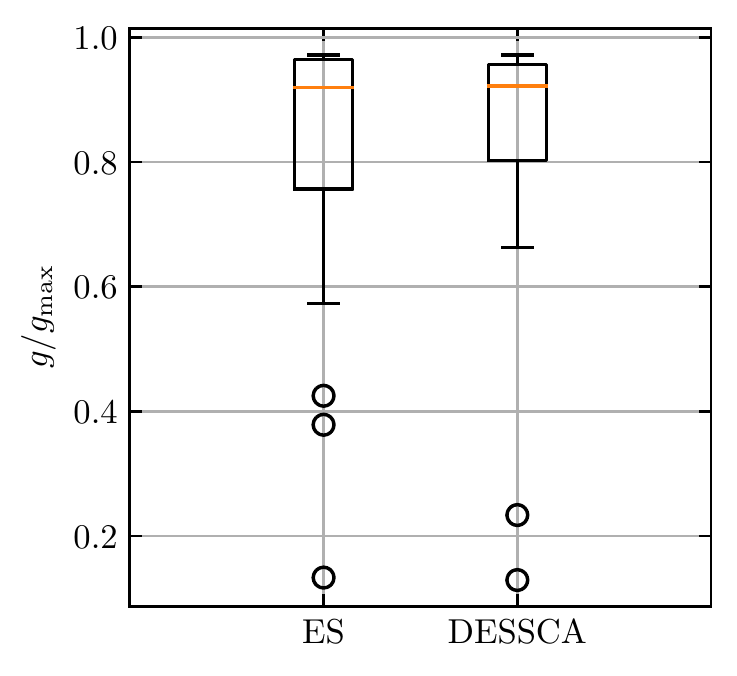}
\subcaption{Mountain car}
\label{fig:MC_Boxplots}
\end{subfigure}
\begin{subfigure}[c]{0.33\textwidth}
\includegraphics[width=1\textwidth]{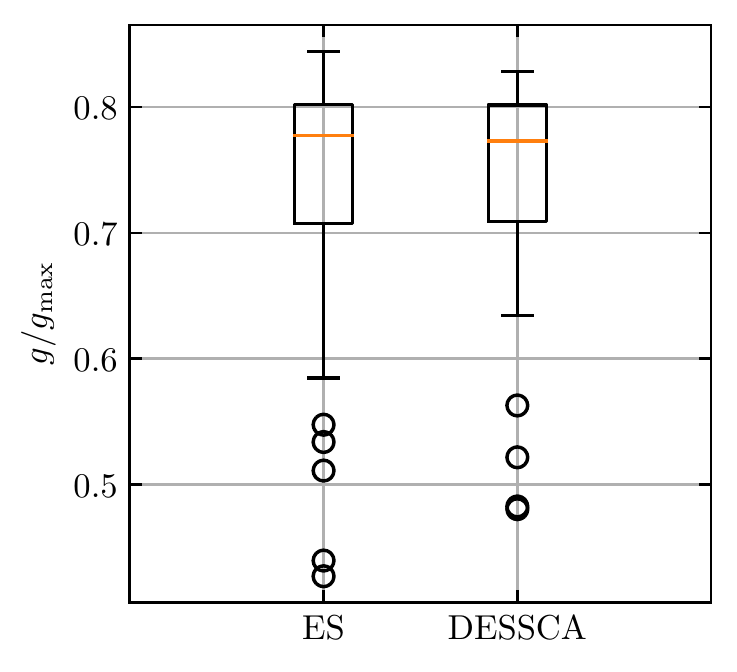}
\subcaption{Cartpole}
\label{fig:CP_Boxplots}
\end{subfigure}
\begin{subfigure}[c]{0.33\textwidth}
\includegraphics[width=1\textwidth]{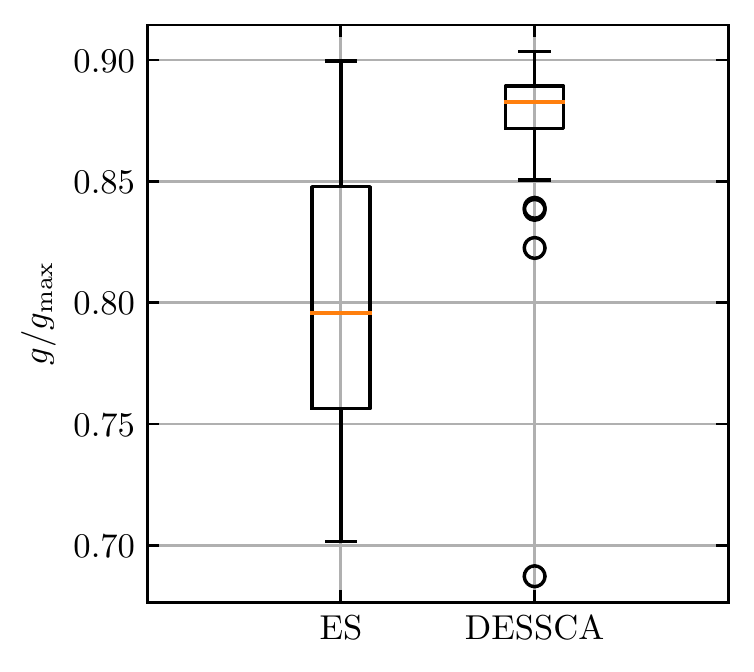}
\subcaption{PMSM current control}
\label{fig:CC_Boxplots}
\end{subfigure}
\caption{Boxplot diagrams showing the performance scattering of 50 independently trained RL control agents based on separate test episode(s)  using different initialization strategies during training.The vertical axis has been normalized with respect to the maximum achievable return, cf. \eqref{eq:performance_normal}.}
\label{fig:Boxplots}
\end{figure*}

The found results presented in Fig. \ref{fig:CC_Boxplots} show that the usage of DESSCA is a major improvement to the yielded performance. The median performance improved by almost $11 \;\%$ while the interquartile range was narrowed down by a large extent and to a much better performance interval. 

\section{Result Discussion}
\label{sec:result}
Overall, the proposed DESSCA algorithm supports the RL training process by improved exploration. In order to evaluate the significance of the collected validation data, a further investigation on the confidence interval of the performance mean is conducted as presented in Tab. \ref{tab:results}. It is assumed that a significant improvement is achieved if the DESSCA $95\;\%$ confidence interval is strictly better than the ES $95\;\%$ confidence interval. As can be found, this significant improvement holds true for all three considered scenarios. Please note that significance does not imply a strong improvement. It can only be inferred that a systematically better performance was achieved in each case. This conclusion based on purely statistical considerations seems a little too benevolent in the cartpole environment, since the boxplots in Fig. \ref{fig:CP_Boxplots} are rather similar. In the motor control scenario (cf. Fig. \ref{fig:CC_Boxplots}), however, the results are very convincing, not only statistically but also in terms of the magnitude of improvement. It is expected, but cannot be proven at this point, that this is due to the earlier mentioned reference tracking task with dynamically changing references signals. Therefore, DESSCA is able to improve the training capability of the considered RL agents  and should be taken into account for future control applications that are to be solved with RL.

\begin{table*}[htb]
    \centering
    \begin{tabular}{l r r r r r r}
    \toprule
    & \multicolumn{2}{c}{\textbf{mountain car}}
    & \multicolumn{2}{c}{\textbf{cartpole} }
    & \multicolumn{2}{c}{\textbf{PMSM current control}}\\
    \midrule

    state space 
    & \multicolumn{2}{c}{continuous, 2 dim.} 
    & \multicolumn{2}{c}{continuous, 4 dim.} 
    & \multicolumn{2}{c}{continuous, 6 dim.} \\

    action space 
    & \multicolumn{2}{c}{continuous, 1 dim.} 
    & \multicolumn{2}{c}{finite, 2 actions} 
    & \multicolumn{2}{c}{continuous, 2 dim.} \\

    algorithm 
    & \multicolumn{2}{c}{DDPG} 
    & \multicolumn{2}{c}{DQN} 
    & \multicolumn{2}{c}{DDPG}\\
    \multicolumn{1}{l}{\multirow{2}{*}{agent's architecture}}
    & \multicolumn{2}{l}{critic: 3 layers with 32 neurons each} 
    & \multicolumn{2}{l}{\multirow{2}{*}{3 layers with 200 neurons each}} 
    & \multicolumn{2}{l}{critic: 4 layers with 128 neurons each}\\
    & \multicolumn{2}{l}{actor: 2 layers with 16 neurons each} 
    & & 
    & \multicolumn{2}{l}{actor: 3 layers with 64 neurons each} \\

    init. strategy 
    & \multicolumn{1}{c}{ES} & \multicolumn{1}{c}{DESSCA} 
    & \multicolumn{1}{c}{ES} & \multicolumn{1}{c}{DESSCA} 
    & \multicolumn{1}{c}{ES} & \multicolumn{1}{c}{DESSCA}\\
    \multicolumn{1}{l}{\multirow{2}{*}{median performance}}
    & \multicolumn{1}{c}{\multirow{2}{*}{$91.941 \; \%$}} 
    & \multicolumn{1}{c}{\multirow{2}{*}{$92.206 \; \%$}} 
    & \multicolumn{1}{c}{\multirow{2}{*}{$77.738 \; \%$}} 
    & \multicolumn{1}{c}{\multirow{2}{*}{$77.308 \; \%$}} 
    & \multicolumn{1}{c}{\multirow{2}{*}{$79.565 \; \%$}} 
    & \multicolumn{1}{c}{\multirow{2}{*}{$88.273 \; \%$}} \\
    & & & & & & \\

    relative median 
    & \multicolumn{2}{c}{\multirow{2}{*}{$+0.289 \; \%$}} 
    & \multicolumn{2}{c}{\multirow{2}{*}{$-0.553 \; \%$}} 
    & \multicolumn{2}{c}{\multirow{2}{*}{$+10.944 \; \%$}}\\
    improvement & & & & & &\\

    \multicolumn{1}{l}{\multirow{2}{*}{interquartile range}}
    & \multicolumn{1}{c}{\multirow{2}{*}{$20.778 \; \%$}} 
    & \multicolumn{1}{c}{\multirow{2}{*}{$15.427 \; \%$}} 
    & \multicolumn{1}{c}{\multirow{2}{*}{$9.451 \; \%$}} 
    & \multicolumn{1}{c}{\multirow{2}{*}{$9.254 \; \%$}} 
    & \multicolumn{1}{c}{\multirow{2}{*}{$9.154 \; \%$}} 
    & \multicolumn{1}{c}{\multirow{2}{*}{$1.734 \; \%$}}\\
    & & & & & &\\

    sample mean 
    & \multicolumn{1}{c}{$83.327 \; \%$}
    & \multicolumn{1}{c}{$85.934 \; \%$}
    & \multicolumn{1}{c}{$73.657 \; \%$}
    & \multicolumn{1}{c}{$74.061 \; \%$}
    & \multicolumn{1}{c}{$79.766 \; \%$}
    & \multicolumn{1}{c}{$87.456 \; \%$}
    \\

    $95 \;\%$ confidence 
    & \multicolumn{1}{c}{\multirow{2}{*}{\shortstack[l]{$[83.015 \;\%,$ \\ \quad $83.638\;\%]$}}}
    & \multicolumn{1}{c}{\multirow{2}{*}{\shortstack[l]{$[85.651 \;\%,$ \\ \quad $86.217\;\%]$}}}
    & \multicolumn{1}{c}{\multirow{2}{*}{\shortstack[l]{$[73.476 \;\%,$ \\ \quad $73.837 \;\%]$}}}
    & \multicolumn{1}{c}{\multirow{2}{*}{\shortstack[l]{$[73.888\;\%,$ \\ \quad $74.233 \;\%]$}}}
    & \multicolumn{1}{c}{\multirow{2}{*}{\shortstack[l]{$[78.234 \;\%,$ \\ \quad $81.299\;\%]$}}}
    & \multicolumn{1}{c}{\multirow{2}{*}{\shortstack[l]{$[86.587 \;\%,$ \\ \quad $88.325 \;\%]$}}}
    \\
    interval of the mean & & & & & &\\

    significant 
    & \multicolumn{2}{c}{yes} 
    & \multicolumn{2}{c}{yes} 
    & \multicolumn{2}{c}{yes}\\
    \bottomrule
    \end{tabular}
    \caption{Configuration and evaluation of the three considered scenarios with respect to the results depicted in \mbox{Fig. \ref{fig:Boxplots}}}
    \label{tab:results}
\end{table*}

\section{Conclusion and Outlook}
\label{sec:outlook}
In the considered application scenarios, the presented DESSCA approach was able to reduce the control performance scatter during test compared to standard exploring starts leading to more consistent control results. Moreover, in two out of three cases the median control performance has been increased. Especially in the motor control scenario, which is a tracking control problem, the DESSCA algorithm achieved a major performance improvement. The suggested algorithm uses the rather simple and well-known methods of KDE and PSO which are available in most programming languages and should allow for a simple and fast implementation within any RL framework. Especially in the context of complicated experiments a proper initialization strategy can help to make efficient use of the training time as it was shown in this contribution.

Despite the concise and simple DESSCA algorithm there remains a large scope for future research. Some interesting questions are listed in the following. Firstly, it could be expected that the DESSCA approach is not only helpful in the context of exploration during episode starts but also feasible for supporting online exploration during an episode, e.g., by allowing expedient changes to the reference state at run time or by guiding (pseudo-)random action selection towards rarely used actions. Furthermore, the observed performance improvements must be examined and verified on additional and varying plant systems with different control quality demands in order to derive general statements on the impact of the initialization strategy.

Moreover, DESSCA can be used not only for exploration but also in the course of testing. Compared to standard design-of-experiment methods \cite{eriksson2000design}, such as Latin hypercube sampling, no discretization of the state space is required and sample points can be generated online step by step. 

Of course, also the selected methods to implement the algorithm are of interest. Among others, it needs to be analyzed whether different tools can fill in the role of KDE in this framework and, lastly, which optimizer is a sensible choice. As both, the KDE as well as the optimizer have their own configuration parameters there are still many degrees of freedom that can be considered to further improve the impact of DESSCA. Also, KDE is a costly operation in terms of computation time if many high-dimensional samples are to be handled. A recursive approach for the estimation of $\hat{c}(\bm{x})$ would therefore be of interest to facilitate the usability of DESSCA in corresponding environments.

Lastly, it was assumed in this contribution that the feasible state space that is utilized during operation is known in beforehand, which is quite sensible in the context of safety critical technical applications. If, however, state constraints can not be defined in beforehand it could be valuable to investigate on methods that can adapt the reference coverage $c^*(\bm{x})$ in terms of unreachable state regions.

\ifCLASSOPTIONcaptionsoff
  \newpage
\fi

\bibliographystyle{IEEEtran}
\bibliography{Sources}

\allowdisplaybreaks
\appendices
\section{Specification of the Considered Environments}
\subsection{Mountain Car}
The normalized state space is defined by
\begin{subequations}
\begin{align*}
        \bm{x}_k &= \begin{bmatrix}
            p_k \\ v_k
        \end{bmatrix}
        \in
        \underbrace{\begin{bmatrix}
            [-1.2, 0.6] 
            \\ [-0.07, 0.07]
        \end{bmatrix}}_{\mathcal{B}},
\end{align*}
with position $p$ and velocity $v$. The continuous action space is scalar and corresponds to the drive force of the car. Furthermore, it must satisfy the box constraint
\begin{align*}
        u_k&\in[-1,1].
\end{align*}
The state update equations are defined by:
\begin{align*}
        v_{k+1}&= \left\{\begin{matrix}
            0, \text{  if } p_k = -1.2 \text{ and } v_k < 0,\\
            v_k + 1.5 \cdot 10^{-3} u_k - 2.5 \cdot 10^{-3} \cos(3 p_k), \text{  else},
        \end{matrix}\right.\\
        p_{k+1}&= p_k + v_{k+1}, 
\end{align*}
which includes that the mountain car can bump into a wall, stopping it immediately. The reward function
\begin{align*}
        r(\bm{x}_k, u_k)&=\left\{\begin{matrix}
            100, & \text{if } p_k > 0.45,\\
            -\frac{1}{10} u_k^2, & \text{else},
        \end{matrix}\right.
\end{align*}
features the aspiration of the target state with a high reward and the penalization of the control effort $u^2$. Lastly, the termination flag
\begin{align*}
    d(\bm{x}_k, u_k)&=\left\{\begin{matrix}
            1, & \text{if } p_k > 0.45,\\
            0, & \text{else},
    \end{matrix}\right.
\end{align*}
defines that system termination occurs after successfully reaching the goal state.
The permitted initialization subspace includes the whole state space in this environment:
\begin{align*}
    \bm{x}_0 &\in \mathcal{B}.
    \label{eq:MC_init}
\end{align*}
\label{eq:MC}
\end{subequations}

\subsection{Cartpole}
The normalized state space is defined by
\begin{subequations}
\begin{align*}
        \bm{x}_k &= \begin{bmatrix}
            p_k \\ v_k \\ \varepsilon_k \\ \omega_k
        \end{bmatrix}
        \in
        \underbrace{
        \begin{bmatrix}
            [-1,1] \\ 
            [-7,7] \\
            [-\pi,\pi] \\
            [-10,10]
        \end{bmatrix}}_{\mathcal{B}},
\end{align*}
with the cart's position $p$ and velocity $v$ and the pole's angular position $\varepsilon$ and angular velocity $\omega$. The discrete action space is a set consisting of two elements
\begin{align*}
    u_k&=\{-1, 1\},
\end{align*}
which relates to the drive force of the cart, and the state update is defined by
\begin{align*}
        b_k&=\frac{10 \, u_k+\omega_k^2 \sin(\varepsilon_k)}{22},\\
        \alpha_k &= \frac{10 \, \sin(\varepsilon_k)- b_k \cos(\varepsilon_k)}{
            \frac{2}{3} - \frac{1}{22} \cos(\varepsilon_k)^2},\\
        a_k &= b_k - \frac{\alpha_k \cos(\varepsilon_k)}{22},\\
        p_{k+1}&= p_k + 0.02 v_k,\\
        v_{k+1}&= v_k + 0.02 a_k,\\
        \varepsilon_{k+1}&= \left(\varepsilon_k + 0.02 \omega_k + \pi\right) \text{mod}\left(2\pi\right) - \pi, \\
        \omega_{k+1}&= \omega_k + 0.02 \alpha_k ,
\end{align*}
wherein the auxiliary variables $a$ and $\alpha$ can be interpreted as the cart's acceleration and the pole's angular acceleration, respectively. Note that $\varepsilon_k \in [-\pi, \pi] \, \forall \, k$. The reward design
\begin{align*}
        r(\bm{x}_k, u_k)&=\left\{\begin{matrix}
            -1, & \text{if } \bm{x}_k \not\in \mathcal{B},\\
            1-|\frac{\varepsilon_k}{\pi}|, & \text{else},\\
        \end{matrix}\right.
\end{align*}
features the targeted goal of stabilizing the pendulum in the upper equilibrium (where $\varepsilon=0$) and penalizes violation of the state limitations. Accordingly, the termination flag
\begin{align*}
        d(\bm{x}_k, u_k)&=\left\{\begin{matrix}
            1, & \text{if } \bm{x}_k \not\in \mathcal{B},\\
            0, & \text{else},
        \end{matrix}\right.
\end{align*}
will be set when such a violation happens. The permitted initialization range is a subset of the state space in which the control agent still has a chance to prevent constraint violations:
\begin{align*}
\bm{x}_0 \in \mathcal{B},
\end{align*}
with further limitations
\begin{align*}
\begin{split}
&\omega_0 \in [-1,1],\\
    &v_0 \left\{\begin{matrix}
    < \sqrt{7.5 (1-p_0)} &\text{if } v_0 > 0.\\
    > -\sqrt{7.5 (1+p_0)} &\text{if } v_0 < 0.\\ 
    \end{matrix}\right.
\end{split}
\end{align*}
\label{eq:CP}
\end{subequations}

\subsection{Permanent Magnet Synchronous Motor Current Control}
The state space of the PMSM in the current control scenario is given by
\begin{subequations}
\begin{align*}
        \bm{x}_k &= \begin{bmatrix}
            i_{\text{d},k} \\ 
            i_{\text{q},k} \\ 
            \omega_{\text{me},k} \\ 
            \varepsilon_{\text{el},k} \\ 
            i_{\text{d},k}^* \\ 
            i_{\text{q},k}^*
        \end{bmatrix}\in
        \underbrace{
        \left\{
        \left.
        \begin{bmatrix}
            [-i_\text{lim}, i_\text{lim}]\\
            [-i_\text{lim}, i_\text{lim}]\\
            [-\omega_\text{lim}, \omega_\text{lim}]\\
            [-\pi, \pi]\\
            [-i_\text{lim}, i_\text{lim}]\\
            [-i_\text{lim}, i_\text{lim}]\\
        \end{bmatrix}\right|
        \begin{matrix}
            i_{\text{s},k} < i_\text{lim}
        \end{matrix}
        \right\}}_{\mathcal{B}},
        \\
        i_{\text{s},k}&=\sqrt{i_{\text{d},k}^2+i_{\text{q},k}^2},
\end{align*}
with the d- and q-current components $i_\text{d}, i_\text{q}$, their respective reference values $i_\text{d}^*, i_\text{q}^*$, the mechanical angular velocity $\omega_\text{me}$ and the electric angular position $\varepsilon_\text{el}$. The continuous action space is defined by
\begin{align*}
        \bm{u}_k&\in
        \left\{
        \begin{bmatrix}
            u_{\text{d},k} \\ 
            u_{\text{q},k}
        \end{bmatrix}
        \left|
        \bm{T}_{32,k}
        \begin{bmatrix}
            u_{\text{d},k} \\ u_{\text{q},k}
        \end{bmatrix} \in 
        \begin{bmatrix}
            [-1,1] \\
            [-1,1] \\
            [-1,1] 
        \end{bmatrix}
        \right.
        \right\},
\end{align*}
with
\begin{align*}
\bm{T}_{32,k}=
            \begin{bmatrix}
            \cos(\varepsilon_k) & -\sin(\varepsilon_k)\\
            \cos(\varepsilon_k-\frac{2\pi}{3}) & -\sin(\varepsilon_k-\frac{2\pi}{3})\\
            \cos(\varepsilon_k-\frac{4\pi}{3}) & -\sin(\varepsilon_k-\frac{4\pi}{3})
        \end{bmatrix},
\end{align*}
which can be interpreted as the duty cycles for the $d$ and $q$ component of the supply voltage which are subjected to the limitations of a usual three-phase two-level voltage source inverter. The state dynamics are defined by the following ordinary differential equations
\begin{align*}
        \frac{\text{d}}{\text{d}t}i_{\text{d}}&=-\frac{ R_\text{s}}{L_\text{d}}i_{\text{d}}+\frac{p \omega_{\text{me}} L_\text{q}}{L_\text{d}}i_{\text{q}}+\frac{U_\text{DC}}{\sqrt{3} L_\text{d}}u_{\text{d}},\\
        \frac{\text{d}}{\text{d}t}i_{\text{q}}&=-\frac{R_\text{s}}{L_\text{q}}i_{\text{q}}\notag
 -\frac{p \omega_{\text{me}} }{L_\text{q}}\left(L_\text{d}i_{\text{d}}+\psi_\text{p}\right)+\frac{  U_\text{DC}}{\sqrt{3} L_\text{q}}u_{\text{q}},\\
        \frac{\text{d}}{\text{d}t}\omega_{\text{me}}&=0, \\
        \frac{\text{d}}{\text{d}t}\varepsilon_{\text{el}}&=p \omega_{\text{me}},\\
        \frac{\text{d}}{\text{d}t}i_{\text{d}}^*&=0,\\
        \frac{\text{d}}{\text{d}t}i_{\text{q}}^*&=0,
\end{align*}
which is dicretized within GEM using the scipy.solve\_ivp integrator from the Scipy library \cite{2020SciPy-NMeth}.
The system parameters are documented in Tab. \ref{tab:pmsm_parameters}. The reward distribution and terminal flag
\begin{align*}
        e_{\text{d/q},k}&=\frac{1}{2}\left(
                    \sqrt{\frac{|i_{\text{d/q},k}^*-i_{\text{d/q},k}|}{2 i_\text{lim}}}+\left(\frac{i_{\text{d/q},k}^*-i_{\text{d/q},k}}{2 i_\text{lim}}
                    \right)^2\right),\\
        r(\bm{x}_k, u_k)&=\left\{\begin{matrix}
            -1, & \text{if } \bm{x}_k \not\in \mathcal{B},\\
            \frac{1-\gamma}{2}\left(1-\frac{i_{\text{s},k}-i_\text{n}}{i_\text{lim}-i_\text{n}}\right)-\frac{1-\gamma}{2}, & \text{if } i_{\text{s},k}>i_\text{lim},\\
            \frac{1-\gamma}{2}\left(
                2-e_{\text{d},k}-e_{\text{q},k}
            \right), & \text{else},\\
        \end{matrix}\right.
\end{align*}
\begin{align*}
        d(\bm{x}_k, u_k)&=\left\{\begin{matrix}
            1, & \text{if } \bm{x}_k \not\in \mathcal{B},\\
            0, & \text{else},
        \end{matrix}\right.
\end{align*}
incorporate 1) a penalty for limit violations and consequent shutdowns of the motor system, 2) a penalty to discourage overcurrent operation, and 3) a reward that increases with decreasing current tracking error $e_{\text{d,q}}$. 
The permitted initialization range is, again, limited to the controllable state subspace of the plant:
\begin{align*}
&\bm{x}_0 \in \mathcal{B},
\end{align*}
\begin{align*}
\begin{split}
\omega_{\text{me},0} \in 
[&
-\omega_\text{me,lim},
\omega_\text{me,lim}
], \quad \varepsilon_{\text{el},0} \in [
-\pi, \pi
],
\\
i_{\text{d},0}, i_{\text{d},0}^* \in 
[&
\text{max}(-i_\text{n}, i_\text{d,c,min}),
\text{min}(i_\text{n},i_\text{d,c,max})
],
\\
i_{\text{q},0}, i_{\text{q},0}^* \in 
[&
\text{max}(-\sqrt{i_\text{n}^2 - i_{\text{d},0}^2}, i_\text{q,c,min}),
\\
&\text{min}(\sqrt{i_\text{n}^2 - i_{\text{d},0}^2}, i_\text{q,c,max})
],
\end{split}
\end{align*}
with
\begin{align*}
\begin{split}
i_\text{d,c,max} &= - \frac{\psi_\text{p}}{L_\text{d}} + \frac{U_\text{DC}}{L_\text{d}\sqrt{3}p|\omega_{\text{me},0}|},
\\
i_\text{d,c,min} &= - \frac{\psi_\text{p}}{L_\text{d}} - \frac{U_\text{DC}}{L_\text{d}\sqrt{3}p|\omega_{\text{me},0}|},
\\
i_\text{q,c,max}^{(*)} &= 
\sqrt{
\left(
\frac{U_\text{DC}}{L_\text{q}\sqrt{3}p|\omega_{\text{me},0}|}
\right)^2 - \frac{L_\text{d}^2}{L_\text{q}^2}
\left(
i_{\text{d},0}^{(*)} + \frac{\psi_\text{p}}{L_\text{d}}
\right)^2
},
\\
i_\text{q,c,min}^{(*)} &= -i_\text{q,c,max}^{(*)}.
\end{split}
\end{align*}
\end{subequations}

\begin{table}[htb]
\centering
\begin{tabular}{l l rl}
\toprule
\textbf{symbol} & \textbf{description} & \textbf{value} & \\
\hline
$p$                 & pole-pair number & $3$ & \\
$R_\text{s}$        & stator resistance & $17.932$ & $\text{m}\Omega$ \\
$L_\text{d}$        & d inductance & $0.37$ & $\text{mH}$\\
$L_\text{q}$        & q inductance & $1.2$ & $\text{mH}$\\
$\psi_\text{p}$     & permanent magnet flux & $65.65$ & $\text{mVs}$\\
$i_\text{n}$        & nominal current & $230$ & $\text{A}$\\
$i_\text{lim}$      & maximum current & $270$ & $\text{A}$\\
$U_\text{DC}$       & DC-link voltage & $350$ & $\text{V}$\\
$\omega_\text{me,lim}$ & maximum angular velocity & $1256.64$ & $\frac{1}{\text{s}}$\\
$\tau$        & sampling time& $100$ & $\upmu \text{s}$\\
\bottomrule
\end{tabular}
\caption{Parameterization of the considered drive system}
\label{tab:pmsm_parameters}
\end{table}

\vfill

\end{document}